\documentclass{article}
\usepackage{ijcai21}

\usepackage{times}
\usepackage{soul}
\usepackage{url}
\usepackage[hidelinks]{hyperref}
\usepackage[utf8]{inputenc}
\usepackage[small]{caption}
\usepackage{graphicx}
\usepackage{amsmath}
\usepackage{amsthm}
\usepackage{booktabs}
\usepackage{algorithm}
\usepackage{algorithmic}
 \usepackage{amssymb}
 \usepackage{breqn}
\usepackage{dsfont}
\usepackage{xcolor}
\usepackage{soul}
\usepackage{epsfig}
\usepackage{graphicx}
\usepackage{amsmath}
\usepackage{amssymb}
\usepackage{booktabs}
\usepackage{dsfont} 
\usepackage{color}
\usepackage{pifont}
\usepackage{graphics}
\usepackage{array}
\usepackage{multirow}
\usepackage{dblfloatfix}
\usepackage{enumitem}
\title{Learn from Anywhere: Rethinking Generalized Zero-Shot Learning with Limited Supervision}

\author{
Gaurav Bhatt$^1$\and
Shivam Chandhok$^1$\And
Vineeth Balasubramanian$^{1}$\\
\affiliations
$^1$IIT Hyderabad\\
\emails
\{gauravbhatt, vineethnb, chandhokshivam\}@iith.ac.in
}
\pubnote{Accepted at IJCAI'21 workshop on Weakly Supervised Representation Learning}

\begin{document}

\maketitle

\begin{abstract}
  A common problem with most zero and few-shot learning approaches is they suffer from bias towards seen classes resulting in sub-optimal performance. Existing efforts aim to utilise unlabeled images from unseen classes (i.e transductive zero-shot) during training to enable generalization. However this limits their use in practical scenarios where data from  target unseen classes is unavailable or infeasible to collect. In this work, we present a practical setting of inductive zero and few-shot learning, where unlabeled images from other \emph{out-of-data} classes, that do not belong to seen or unseen categories, can be used to improve generalization in any-shot learning. We leverage a formulation based on product-of-experts and introduce a new AUD module that enables us to use unlabeled samples from \emph{out-of-data} classes which are usually easily available and practically entail no annotation cost. In addition, we also demonstrate the applicability of our model to address a more practical and challenging,  Generalized Zero-shot under limited supervision setting, where even base seen classes  do  not  have  sufficient  annotated  samples.
We evaluate the proposed method's performance on several established benchmark datasets - CUB, SUN, AWA1, and AWA2, and show that our proposed approach enhances performance on several datasets. To show the proposed method's scalability, we also present experiments on the ImageNet dataset. Furthermore, when there is limited supervision in such settings, the proposed training paradigm outperforms current state-of-the-art techniques.\\
\end{abstract}

\section{Introduction}
\label{sec_intro}
\vspace{-4pt}
Classifying visual concepts for classes which are not available during training has been one of the prominent yet open problems in machine learning. Zero-shot learning (ZSL) aims to tackle this problem where the model has access to a set of seen classes, and the objective is to leverage semantic information (in the form of word or attribute embeddings) to learn visual-semantic relationships which enables generalization to  unseen classes at test-time \cite{devise,good_bad,revise,over_complete_CVPR20,hyperbolic_CVPR20}.\par
Based on images available to the model during training, zero-shot learning can be categorized into two settings: \textit{Inductive} and \textit{Transductive} ZSL. In inductive ZSL, model has access to labeled image-semantic embedding pairs for seen classes only. On the other hand, transductive ZSL  refers to the setting where in addition to the labeled seen class data, we also have access to unlabeled samples from target unseen classes during training. 
The assumption of presence of images from unseen target classes is subject to availability and feasibility. Furthermore, assuming the presence of unseen class samples during training restricts the applicability of the model in practical scenarios where images from specific target classes might not be available. We hence focus on the more challenging inductive setting in this work. However, usually, unlabeled images, which may neither belong to seen or target unseen classes but are available publicly, are abundantly available in practical scenarios. We call these images \emph{out-of-data} samples since they may not necessarily belong to the given dataset. For e.g., given a dataset of birds, the \emph{out-of-data} class samples (unlabeled) can belong to any OpenImage/ImageNet dataset classes other than the seen or unseen classes present in the birds dataset (to avoid any additional information on the classes being studied, for fair evaluation). \par 
To leverage the abundance of these unlabeled \emph{out-of-data} samples, we propose a new 'Learn from Anywhere' paradigm where the model can utilize unlabeled samples from outside the dataset (in particular, from classes outside the dataset) under consideration. Note that this still falls under the inductive zero-shot setting as we only assume presence of images from seen or \emph{out-of-data} classes and not unseen target classes.
Formally,  we propose a new methodology that can leverage unlabeled data from seen or \emph{out-of-data} classes, enabling us to ``learn from anywhere'' and enhance performance of generalized zero/few shot settings. The introduced \emph{out-of-data} samples act as a regularizer enabling the model to learn image structure and visual-semantic relationships which help alleviate bias towards seen classes, resulting in better generalization at test-time. Note that we refer to these unlabeled \emph{out-of-data} samples as AUD (auxiliary data) henceforth. We formulate our model to deal with the scenario where specific modalities (e.g word or attribute embeddings) from AUD samples may be missing, allowing our methodology to be robust/useful in scenarios where paired multimodal data is only partially available.

In addition to results on generalized zero-shot learning (GZSL) and few-shot learning, our use of AUD allows us to demonstrate the applicability of our proposed method for generalized zero-shot in a limited supervision setting where even base seen classes do not have sufficient labeled examples. We refer to this setting as Generalized Zero-Shot Learning with Limited Supervision, henceforth.
This setting is in contrast to existing work on zero-shot or few-shot learning, which assume that there are sufficient annotated examples in the base seen classes \cite{cada,f-vaegan-d2,over_complete_CVPR20,revise,good_bad} or do not have a mechanism to leverage unlabeled or unpaired (missing semantic modality) seen class samples \cite{verma2020meta} to improve performance.\par
Our overall setup is closer to more practical real-world situations where we may not have large numbers of labeled seen class samples or paired multimodal data. This makes us unique compared to current methods \cite{cada} \cite{verma2020meta} which cannot handle such scenarios. Finally, to have a fair GZSL evaluation, we ensure none of the data related to unseen target classes are present in the AUD samples in our experiments.
Our key contributions are summarized as follows: 
\vspace{-3pt}
\begin{itemize}[leftmargin=*]
\setlength\itemsep{-0.2em}
\item We introduce a new `Learn from Anywhere' paradigm and propose a methodology based on Product-of-Experts (POE) formulation to improve zero/few-shot learning. We introduce a new AUD module in this framework that allows us to utilize unlabeled data from seen or \emph{out-of-data} classes during training.
    
    
    
    \item The newly introduced AUD module shares weights with the POE model and improves visual and semantic alignment across the data involved. This helps improve performance by regularizing the model and alleviating bias towards seen classes, allowing our methodology to be helpful in the presence of \emph{out-of-data} samples and `GZSL with Limited Supervision' (few annotated base class samples) settings.
    
    \item We show that the proposed model enhances the performance on generalized zero and few-shot learning when evaluated on several benchmark datasets: CUB, SUN, AWA1, and AWA2.
    
    \item We also demonstrate the the model can better tackle limited supervision in generalized zero-shot setup than several SOTA methods and is robust under scenarios where paired multimodal samples are not available during training (missing modality problem). 
    
\end{itemize}
To the best of our knowledge, this is the first effort that aims to leverage abundantly available unlabeled \emph{out-of-data}-classes/samples which belong neither to seen nor unseen classes to improve inductive generalized zero-shot and few-shot recognition performance. Going beyond existing efforts \cite{verma2020meta}, our method is also novel in allowing the use of unlabeled seen class data for inductive GZSL under limited supervision as well as handling missing modalities during training. 

\section{Related Work}
\label{sec_related_work}
\vspace{-4pt}

Zero-shot learning (ZSL) is a classification problem where the label space is divided into two sets of categories: seen and unseen/novel classes \cite{devise,revise,good_bad,latem,f-clswgan,cada,f-vaegan-d2}. To enable models to classify even unseen classes, training samples typically consist of auxiliary information such as attribute embeddings that bridge the semantic gap between seen and unseen classes. A variant of ZSL, which is relatively less hard, is few-shot learning (FSL), where the training procedure has access to some labeled data from each unseen class \cite{good_bad,cada,f-vaegan-d2,f-clswgan}. Generalized zero and few-shot learning is a practical variant, where the performance evaluation is performed on both seen and unseen classes at test time.  



Recently, researchers have achieved success through the use of generative models  \cite{cada,f-vaegan-d2,verma2017simple,f-clswgan} and statistical methods \cite{sync,eszsl} for any-shot learning.
In a recent state-of-the-art approach, \cite{cada} used Variational Autoencoders (VAEs) to increase the cross-alignment between visual features and semantic embeddings. In addition to this, \cite{dascn,gdan,taco} propose dual adversarial learning paradigms to model visual-semantic joint and enhance knowledge transfer between visual and semantic spaces.
 To deal with the problem of bias towards seen classes in zero-shot learning, researchers have introduced adversarial sampling \cite{dual_adv}, embedding models \cite{corep} or leveraging unlabeled data \cite{revise,protonet,f-vaegan-d2,trans_prop,verma2020meta}. However, none of the existing methods focus on using \emph{out-of-data}-samples or data samples with a missing modality, which we focus on in this work.

\textbf{Relationship to previously proposed methods}. 
The efforts closest to our proposed approach are CADA-VAE \cite{cada}, Meta-ZSL \cite{verma2020meta}, JMVAE \cite{jmvae} and MVAE \cite{mvae}, each in different ways. There are however fundamental differences.
Firstly, we introduce the AUD module, which allows us to use unlabeled \emph{out-of-data} classes/samples for zero-shot recognition during training, which is not possible with any of the methods mentioned above. Similar to us, the CADA-VAE model \cite{cada} uses VAEs to transform the visual features and attribute embedding to latent spaces. However, it relies on the alignment of data from different modalities to compute the joint space. 
This alignment method fails when one of the modalities is missing since all modalities are required during training. In contrast, we design our methodology so that we can seamlessly work under this scenario; the use of a POE network and AUD module in our method helps us model the joint distribution under such settings and use unlabeled data with missing modalities to improve performance. 
Meta-ZSL \cite{verma2020meta} study their approach under limited supervision but cannot handle the missing modality scenario either (and hence cannot take advantage of the availability of unpaired data ).

\begin{figure*}[t!]
\centering 
\includegraphics[scale=0.7]{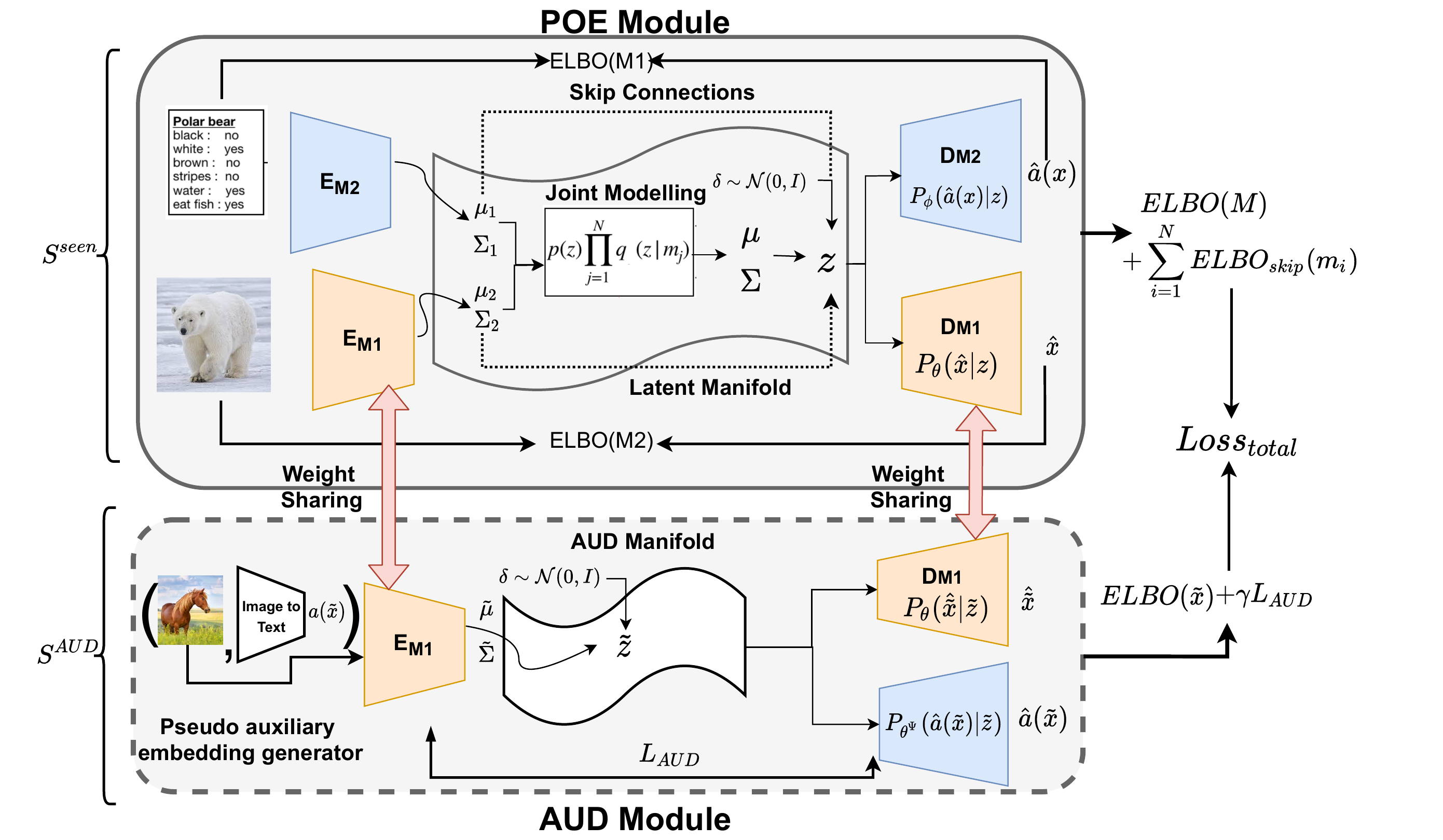}
\caption{
Overview of proposed methodology. Our pipeline comprises of a POE Module (top) and an AUD Module (bottom) as shown. Paired example from $S^{seen}$ includes the image of a \textit{polar bear} along with its corresponding attribute vector. Here, the \textit{horse} image (from ImageNet) depicts a random unlabeled  $S^{AUD}$ sample. 
}
\vspace{-15pt}
\label{fig:one}
\end{figure*}
MVAE \cite{mvae} and JMVAE \cite{jmvae} also use POE networks, similar to our approach, but can leverage unlabeled data only during inference (we use it during the training phase) and also do not address the generalized zero-shot setting. In terms of architecture, we further introduce the idea of using skip connections in our generative network to improve the proposed model's feature generation capability, which is an improvement over MVAE and JMVAE architectures. In this work, we also note that both the learn-from-anywhere and GZSL-with-limited-supervision settings are studied on the more challenging inductive setting, where the model does not have access to any data samples of unseen classes at training time. We now present our methodology.




\vspace{-10pt}
\section{Methodology}
\label{sec_method}
\vspace{-4pt}
\noindent \textbf{Overview:}
Existing methods operate on regimes where model has access to labeled image-semantic embedding pairs from seen classes i.e inductive zero-shot setting or where samples during training also include unlabeled target unseen class images i.e transductive zero-shot setting. However, they are not designed to leverage unlabeled samples from \emph{out-of-data} classes (i.e neither seen nor unseen categories) which are usually present in abundance and accessable with minimal effort and practically no annotation cost. We conjecture that these unlabeled \emph{out-of-data} samples can act as a regularizer enabling the model to better learn visual-semantic structure and aid generalization to novel classes by alleviating bias in limited supervision settings.\\
As shown in Figure \ref{fig:one}, we propose a methodology based on product of experts formulation which enables us to utilise unlabeled \emph{out-of-data} samples to enhance generalization. Specifically, we introduce an AUD module (which shares weights with the POE modules) to incorporate the \emph{out-of-data} samples in modelling the image-semantic joint ( As explained in Section \ref{learn-from-anywhere} ). We formulate the AUD module such that it can effectively incorporate samples with missing modalities enabling  our model to be robust/useful in scenarios where such paired multimodal data is only partially available.\\ 
In addition to this, we show that the proposed approach can be leveraged to address a more practical and difficult in GZSL with Limited Supervision setting where even base seen classes do not have sufficient annotated examples (As discussed in Section \ref{gzsl_lim_sup})\\
\noindent \textbf{Problem Setting:} 
Let $S^{seen}=\{(x, y, a(y))|x \in X, y \in Y^{seen}, a(y) \in A \}$  be the set of seen class data; where $X$ corresponds to set of image features (extracted from a pre-trained model), $Y^{seen}$ corresponds to the set of seen class labels, and $A$ denotes the set of corresponding attributes\cite{good_bad}. Similarly, the set for novel/unseen classes are defined as $S^{novel}=\{(n,a(n))|n \in Y^{novel}, a(n) \in A\}$, where $Y^{novel}$ corresponds to the set of novel/unseen class labels.  Note that attribute information is provided for these novel classes, but no image data is available. In GZSL, the objective is to learn a classifier that can classify both seen and unseen class images at test time.
In our work, we further divide the data samples as annotated and unlabeled. Let the annotated samples for the seen classes be given as $S^{seen}$ (defined previously), and the auxiliary unlabeled data (AUD) is given by $S^{AUD}=\{\Tilde{x}| \Tilde{x} \in \Tilde{X} \}$. 

We finally address the GZSL task using $S^{lim\_sup} = \{S^{seen} \bigcup S^{AUD}\}$ to train our model, while the final performance is studied on a test set that includes class labels from $Y^{seen} \bigcup Y^{novel}$.The samples in the set $S^{AUD}$ vary according to the settings we address. In the 'Learn from Anywhere' paradigm, $S^{AUD}$ contains unlabeled samples from \emph{out-of-data} classes. In the 'GZSL with Limited Supervision' setting $S^{AUD}$ contains unlabeled seen class samples. We provide detailed information about the setup in the respective sections for each setting.
.
\vspace{-5pt}
\subsection{Prelimnaries}
\vspace{-2pt}


\noindent \textbf{Variational autoencoders}.
We use a standard VAE \cite{vae}, a latent variable model that tries to find the true conditional distribution over the latent variables. We use $z$ to denote a common latent variable that is conditioned on seen annotated data pairs. The VAE takes the form $p_\theta (x,z) = p_\theta(z) p_\theta(x|z)$, where $p_\theta(z)$ is the prior distribution (typically assumed to be a standard normal $\mathcal{N}(0,1)$), $p_\theta(x|z)$ is a decoder network, parametrized by $\theta$, which generates $x$ given $z$. To approximate the true posterior, we fit an inference network of the form $q_\phi(z|x)$. The inference network (or the encoder) predicts values for $\mu$ and $\Sigma$ such that $q_\phi(z|x)$ = $\mathcal{N}(\mu,\Sigma)$. The latent variable $z$ is sampled using the reparamaterization trick \cite{vae}. In our work, we slightly modify this reparametrization so that it better aligns with our training objective (discussed later in this section \ref{learn-from-anywhere}). 

The loss function for a VAE is the variational bound on the marginal likelihood (the evidence lower bound, ELBO) and can be computed for a single data point as: 
\vspace{-3pt}
\begin{dmath}
ELBO(X) =  {E_{q_\phi(z|x)}}\ [\log\ p_\theta(x|z)]-\beta\ D_{KL}(q_\phi(z|x) || p_\theta (z)),
\end{dmath}
\vspace{-2pt}
where $\beta$ is the annealing term that lets VAE learn ``important'' representations before they are ``smoothed out'' \cite{cada}, and $D_{KL}$ is the Kullback-Leibler (KL) divergence.

For zero-shot learning, we condition the VAEs on the image features and corresponding attribute embedding. In an ideal case, one can incorporate as many additional meta inputs (e.g., sentence encoding, word vector embedding corresponding to class labels) that may be available in a given context. In that case, the ELBO loss for a single data sample with multiple modalities like image, attributes, and auxiliary word vectors, i.e. $M = \{ x, a(y), w(y), ... \}$ (where $a(y)$, $w(y)$ are different kinds of auxiliary information that may be available) can be given by:
\vspace{-3pt}
\begin{dmath}
ELBO(M) =  { E_{q_(z|M)}}\ \left[ \sum_{m_i \in M} \log\ p_\theta(m^i|z)\right]-\beta\ D_{KL}(q_\phi(z|M) || p_\theta (z)).
\label{eqn_elbo_m}
\end{dmath}
\subsection{Learn from Anywhere}
\label{learn-from-anywhere}
\vspace{-2pt}
This section formally introduces our model components and describes how the proposed methodology can utilize unlabeled samples from \emph{out-of-data} classes to improve zero and few-shot performance.
In this setting, the set $S^{AUD}$ contains unlabeled multimodal samples from \emph{out-of-data} classes (which belong neither to the seen classes nor to unseen categories of the dataset as mentioned before). Furthermore, we also show how our model can work even when specific modalities (e.g semantic attributes/embeddings) from samples are missing.

The proposed architecture is shown in Figure \ref{fig:one}, and the training procedure is outlined in Algorithm \ref{alg:algorithm}. We now present the proposed method.

\noindent \textbf{Learning Joint Distribution with AUDs}. 
The ELBO term defined in Equation \ref{eqn_elbo_m} depends on an underlying assumption that all training and testing samples have paired information provided, i.e., every image feature should have a corresponding attribute embedding. However, with unlabeled missing modality data (where there are only images), we do not have access to the meta-information (such as attributes).
A quick fix would be to use 0-vector for attributes corresponding to unlabeled  data, but this can affect the KL-divergence computation. As a remedy, we consider a graphical model where we assume that the data, attributes, and any other auxiliary information present are independent given a common latent representation. 
This assumption helps us model to the joint distribution as:
\vspace{-3pt}
\begin{dmath}
p(z,m_1,...,m_N) =  p(z) p_\theta(m_1|z)....... p_\theta(m_n|z),
\label{eqn_joint}
\end{dmath}
where $\{m_1, ......, m_N\} \sim \{ x, a(y), w(y), ... \}$ are the $n$-modalities and $z$ is the shared latent representation.\\
Note that this factorization (Equation \ref{eqn_joint}) allows us ignore a missing attribute corresponding to unlabeled data while computing the marginal likelihood \cite{mvae,jmvae} and tackle the missing modality problem during training  . We define the inference network on all given modalities as (shown as joint modelling in Figure \ref{fig:one}):
\vspace{-3pt}
\vspace{-3pt}
\begin{dmath}
p(z|m_1,...,m_N) \propto  p(z) \prod_{i=1}^N q(z|m_i),
\end{dmath}
where $p(z)$ is the prior expert and $q(z|M)=\prod_{i=1}^N q(z|m_i)$ is used to approximate the true posterior distribution.  This is also known as product-of-experts (POEs) \cite{poe}. An advantage of using such POE networks is that it has a closed form analytical solution when $p(z)$ and $q(z|X)$ are assumed to be Gaussian. 

The joint modeling network outputs $\mu$ and $\Sigma$, which are used to sample a latent variable $z$. One choice of design to sample $z$ would be to reparameterize each modality, sample a latent variable ($z_i$), and then multiply them independently. 
This design omits the use of joint modelling network and simply relies on the assumption of conditional independence (Equation \ref{eqn_joint}). We conjecture that this design may not be effective as overall noise introduced in the latent codes may destabilize the training (we show the ineffectiveness of this approach in our ablation studies, Section \ref{ablation}). Instead, we compute the joint parameters of the inference network as: $\mu = (\sum_{i}\mu_i\ T_i)(\sum_{i}T_i)^{-1}$ and $\Sigma = (\sum_{i}T_i)^{-1}$, where $\mu_i$ and $\Sigma_i$ are the parameters for the $i^{th}$ expert and $T_i$ is the inverse of covariance $\Sigma_i$.
Finally, we sample a common/global latent variable using the standard reparametrization: $z =  \mu + \delta \odot \Sigma$, with $\delta \sim N(0,1)$ as Gaussian noise \cite{vae}.

\noindent \textbf{Training with auxiliary unlabeled data}.
\label{training_pseudo}
 Methods such as MVAE\cite{mvae} and JMVAE\cite{jmvae} design their inference procedure to handle unlabeled data during testing. However, in this work, we seek to incorporate AUDs during training to minimize the class bias of seen classes in the GZSL setting. Therefore, we formulate a novel training procedure that can deal with AUDs during training and inference. 

\begin{algorithm}[t!]
\footnotesize
\caption{Proposed Training Procedure}
\label{alg:algorithm}
\textbf{Input}: $ x,\ x(a),\ \Tilde{x},\ a(\Tilde{x}),\ \mathds{1}_\alpha$ \\
\textbf{Parameter}: $\phi$, $\theta$, $\theta^\psi$, $\theta^\alpha$ \\
\textbf{Output}: $\mu, \sigma$
\begin{algorithmic}[1] 
\FOR{each sample in the dataset $<X or \Tilde{X}>$}
\FOR{modality \textbf{k} in given data-sample}
\STATE compute $\mu_k$ and $\sigma_k$
\ENDFOR
\STATE initialize p(z) $\sim N(0,1)$ and $\delta \sim\ N(0,1)$
\STATE$\mu$, $\sigma$ =  $\mu_z*\prod_{k} (\mu_k$), $\sigma_z*\prod_{k} (\sigma_k$)
\STATE \textit{z} = $\mu + \delta \odot \sigma$; $\hat{\alpha}$ = $P_{\theta^\alpha}(\hat{\alpha}|z)$
\IF { $\mathds{1}_\alpha == 1$}
\STATE loss = $ ELBO(M) + \sum_{i=1}^N ELBO_{skip}(m_i)$ + $log(\hat{\alpha}$)
\ELSE
\STATE loss = $ELBO(\Tilde{x}) + \gamma\ L_{AUD}$ + (1-$log(\hat{\alpha}$))
\ENDIF
\STATE update $\theta,\ \phi,\ \theta^\psi$ using \textit{Adam} optimizer
\ENDFOR
\STATE \textbf{return} $\mu, \sigma$
\end{algorithmic}
\vspace{-3pt}
\end{algorithm}
\vspace{-3pt}
We start by introducing a binary feature $\mathds{1}_\alpha$ (or indicator variable) that tells us whether a given training sample is paired or missing modality (AUD). This binary feature can be computed offline before training. Each training sample is now given as a triplet $\{x, a(x), \mathds{1}_\alpha(x, a(x))\}$.
For a paired triplet ($\mathds{1}_\alpha$ = 1) we have both the image feature $x$ and attribute $a(x)$. In this case we first sample a global latent variable ($z$) from the joint inference network and compute the likelihood of the image feature ($P_\theta(\hat{x}|z)$) along with the attribute embedding ($P_\theta(\hat{a}(x)|z)$). For AUDs with missing modalities ($\mathds{1}_\alpha$ = 0), we have only the image feature $\Tilde{x}$ and attribute information $a(x)$ is not available.\\ 
Next, we compute the joint ELBO only for triplets where $\mathds{1}_\alpha$ = 1  using the POE network (top part of figure \ref{fig:one}). For AUDs where $\mathds{1}_\alpha$ = 0, we compute the ELBO only on the image feature $x$ (the image encoder and decoder is shared among paired samples and AUDs). This enables our model to utilize both paired and unlabeled missing modality samples.\par

Now that we have described the training procedure in case of AUDs with missing modalities, in order to further improve our model, another addition can be to generate the auxiliary semantic embedding $a(\Tilde{x})$ corresponding to an unlabeled image $\Tilde{x}$. We train an image-to-text generative model to generate a auxiliary semantic embedding corresponding to the image (we use the 512-dimensional output of the penultimate layer of a bottom-up attention network in \cite{botton_up_atten}).We refer to them as pseudo-auxiliary semantic embeddings henceforth.
Note that we make sure that our image-to-text model is not trained on any of the unseen classes, so it does not violate the zero-shot condition. This way, the pseudo-auxiliary semantic embeddings are computed for any unlabeled image without compromising the GZSL setting and are extracted before training just as we compute visual features for images.  

One downside of using additionally incorporating pseudo-auxiliary semantic embeddings is that there is no way we can evaluate their quality without human intervention. Thus, computing the joint distribution using irrelevant pseudo-auxiliary semantic embedding could make the joint representation ill-posed. As a solution, 
a separate decoder ($P_{\theta^{\psi}}(\hat{a}(\Tilde{x})|\Tilde{z})$) (as shown in figure 1) is used to minimize the likelihood loss of pseudo-auxiliary semantic embeddings, where $\Tilde{z}$ is the latent variable conditioned on unlabeled image $\Tilde{x}$ ($\Tilde{z}$ is generated using the common image encoder, as shown in Figure \ref{fig:one}). That is, when generated  pseudo-auxiliary semantic embeddings is given as input, we compute the $L_1$-norm between the generated embedding $\hat{a}(\Tilde{x})$ and the ground truth $a(\Tilde{x})$ obtained from the image-to-text pre-trained model:

\vspace{-3pt}
\begin{dmath}
L_{AUD} =  \underset{\substack{{\Tilde{z} \sim q(\Tilde{z}|\Tilde{x}), \Tilde{x} \sim \Tilde{X}}}} {\mathbb{E}}|| P_{\theta^{\psi}}(\hat{a}(\Tilde{x})|\Tilde{z}) - a(\Tilde{x}) ||_{1}. 
\end{dmath}
\vspace{-4pt}
Using this formulation, we ensure that the latent variable $\Tilde{z}$ is not conditioned on the pseudo-attribute embedding $a(\Tilde{x})$ for AUDs, while the decoder $P_{\theta^{\psi}}(\hat{a}(\Tilde{x})|\Tilde{z})$ learns to map unlabeled images to pseudo-attributes.

\noindent \textbf{Skip connections}. To improve the network latent representation capability, we introduce the skip connections in the proposed architecture. We define a skip connection for modality $m_i$ as the ability to generate itself independently (as shown in Figure \ref{fig:one}). The latent variable $z_i$ is conditioned on modality $m_i$ alone and is sampled using the standard reparameterization trick \cite{vae}.
Using $z_i$ and the skip connection, the loss for modality $m_i$ is given by:
\vspace{-3pt}
\begin{dmath}
ELBO_{skip}(m_i) =  {E_{q_\phi(z_i|m_i)}}\ [ \log\ p_\theta(m_i|z_i)]-\beta\ D_{KL}(q_\phi(z_i|m_i) || p_\theta (z_i)).
\end{dmath}

Our final training objective for a single data point is given by:
\vspace{-5pt}
\begin{dmath}
\mathds{1}_\alpha \bigg ( ELBO(M) + \sum_{i=1}^N ELBO_{skip}(m_i) \bigg ) + (1-\mathds{1}_\alpha) \bigg ( ELBO(\Tilde{x}) + \gamma\ L_{AUD} \bigg )
\end{dmath}
where, $\gamma$ is the pseudo-auxiliary semantic embedding factor which is manually tuned based on the choice of the AUD dataset. 

\begin{table*}[h!]
\footnotesize
\centering
\resizebox{\textwidth}{!}{%
\begin{tabular}{c|ccc|ccc|ccc|ccc}
    \toprule
    
    & \multicolumn{3}{c}{CUB} & \multicolumn{3}{c}{SUN} & \multicolumn{3}{c}{AWA1} & \multicolumn{3}{c}{AWA2} \\
    
    Method & s & u & H & s & u & H & s & u & H & s & u & H \\
    \midrule
    \textbf{DeViSE}(NeurIPS'13) \cite{devise} &53.0 & 23.8 & 32.8 & 27.4 & 16.9 & 20.9 & 68.7 & 13.4 & 22.4 & 74.7 & 17.1 & 27.8
    \\
    
    \textbf{ESZSL}(ICML'15)\cite{eszsl} & 63.8 & 12.6 & 21.0 & 27.9
    & 11.0 & 15.8 & 75.6 & 6.6 & 12.1 & 77.8 & 5.9 & 11.0\\
    
    \textbf{SYNC}(CVPR'16) \cite{sync} & 70.9 & 11.5 & 19.8 & 43.3 & 7.9 & 13.4 & 87.3 & 8.9 & 16.2 & 90.5 & 10.0 & 18.0\\
    
    \textbf{ReViSE}(ICCV'17) \cite{revise} & 28.3 & 37.6 & 32.3 & 20.1 & 24.3 & 22.0 & 37.1 & 46.1 & 41.1 & 39.7 & 46.4 & 42.8\\
    \midrule

    \textbf{SE-GZSL}(CVPR'18)\cite{verma_se}  &53.3 &41.5 &46.7  &30.5 &40.9 &34.9 &67.8 &56.3  &61.5 &68.1 &58.3 &62.8\\
    \textbf{f-CLSWGAN}(CVPR'18) \cite{f-clswgan} & 57.7 & 43.7 & 49.7 & 36.6 & 42.6 & 39.4 & 61.4 & 57.9 & 59.6 & 68.9 & 52.1 & 59.4\\
    
    \textbf{Cyc-WGAN}(ECCV'18) \cite{cycWGAN} &59.3 &47.9 &53.0 &33.8 &47.2  &39.4 &63.4 &59.6 &59.8 &- &- &- \\
    
    \textbf{CVAE-GZSL}(CVPRW'18)\cite{cvae-zsl} &- &- &34.5 &- &- &26.7 &- &- &47.2 &- &- &51.2\\
    
    \textbf{CADA-VAE}(CVPR'19) \cite{cada} & 53.5 & 51.6 & 52.4 & 35.7 & 
    47.2 & 40.6 & 72.8 & 57.3 & 64.1 & 75.0 & 55.8 & 63.9\\
    
    \textbf{f-VAEGAN-D2}(CVPR'19) \cite{f-vaegan-d2} &60.1 & 48.4 & 53.6 & 
    38.0 & 45.1 & 41.3 & 
    70.6 & 57.6 & 63.5 & - & - & -\\
    
    \textbf{DASCN}(NeurIPS'19) \cite{dual_adv} & 45.9 & 59.0 & 51.6 & 
    42.4 & 38.5 & 40.3 & 59.3 & 68.0 & 63.4 & - & - & -\\
    
    \textbf{SGAL}(NeurIPS'19) \cite{simul_gen} & 55.3  & 40.9 & 47.0 & 
    34.4  & 35.5 & 34.9 & 74.0 & 52.7 & 61.5 & 86.2 & 52.5 & 65.3\\

    \textbf{CRnet}(ICML'19) \cite{corep} & 56.8 & 45.5 & 50.5 & 
    36.5 & 34.1 & 35.3 & 74.7 & 58.1 & \underline{\textbf{65.4}} & 78.8 & 52.6 & 63.1\\
    
    \textbf{SGMAL}(NeurIPS'19) \cite{semantic_guide} & 71.3 & 36.7 & 48.5 & 
    - & - & - & 87.1 & 37.6 & 52.5 & - & - & -\\
    
    \textbf{VSE}(CVPR'19)\cite{vse} &68.9&39.5&50.2 &- &- &- &- &- &-  &88.7&45.6& 60.2\\
     
    \textbf{IIR}(ICCV’19) \cite{IIR} &52.3 &55.8  &53.0 &30.4 &47.9 &36.8 &- &- &- &83.2 &48.5  &61.3 \\
    
    \textbf{TCN}(ICCV'19) \cite{TCN} &52.6 &52.0 &52.3 &31.2 &37.3 &34.0 &49.4 &76.5 &60.0 &61.2 &65.8 &63.4 \\
    \textbf{LisGAN}(CVPR'19) \cite{lisGAN} &57.9 &46.5 &51.6  &37.8 &42.9 &40.2 &76.3 &52.6 &62.3 &- &- &- \\
    
    \textbf{SGMA}(NeurIPS’19) \cite{SGMA} &71.3&36.7&48.5 &-&- & - &- &- &-  &-&-& -\\
    
    \textbf{LsrGAN}(ECCV’20) \cite{LSRGAN} &58.1&48.1&53.0 &37.7 &44.8  & 40.9 &- &- &-  &-&-& -\\
    
    \textbf{ZSML}(AAAI'20) \cite{verma2020meta} & 60.0 & 52.1 & 55.7& 
    - & - & - & 57.4 & 71.1 & 63.5 & 58.9 & 74.6 & \underline{\textbf{65.8}}\\
    
    \textbf{OCD-CVAE}(CVPR'20) \cite{over_complete_CVPR20} & 44.8 & 59.9 & 51.3 & 
    44.8 & 42.9 & 43.8 & - & - & - & 59.5 & 73.4 & 65.7\\
   
       \midrule
    
    Proposed (AUD - ImageNet) & 56.8 & 52.1 & 54.3 & 39.7 & 48.9 & \underline{\textbf{43.9}} &
    78.8 & 53.9 & 64.2 & 81.7 & 54.5 & 65.4\\
    
    Proposed  (AUD - OpenImage) & 57.9 & 54.3 & \underline{\textbf{56.1}} & 39.5 & 48.4 & 43.4 &
    78.5 & 55.2 & \textbf{64.9} & 81.8 & 54.7 & \textbf{65.6}\\
    
    \bottomrule
  \end{tabular}
}
\caption{Comparison of proposed architecture with several recent baseline methods and state-of-art methods on Generalized Zero-Shot Learning . We report accuracy ($\%$) of seen and unseen classes (u,s) along with their harmonic mean (H). Note that '-' implies not reported}
\label{tab:gzsl}
\end{table*}

\vspace{-2pt}
\subsection{GZSL with Limited Supervision}
\label{gzsl-lim-sup}
\vspace{-4pt}
In addition to the learn-from-anywhere paradigm, in this section, we discuss the application of our proposed methodology in a more practical, challenging setting where even base seen classes do not have sufficient labeled examples, however, we have access to unlabeled seen class samples i.e GZSL under limited supervision.\\ 
This setting can also be viewed as a combination of zero-shot and semi-supervised learning (SSL), where we have very inadequate annotated samples in base seen classes and other unlabeled seen class samples are available. We refer to these unlabeled seen class samples as missing data samples as their labels are missing.
Note that this is a more difficult setting than the standard zero-shot setting since we are restricting training conditions from two perspectives: no unseen class images (ZSL), as well as inadequate, annotated, seen class samples (SSL).
Also, note that this setting is a special case of 'Learn from Anywhere' paradigm where unlabeled AUD samples also belong to seen classes, instead of \emph{out-of-data} classes (as considered in learn-from-anywhere setting).

The AUD module in our methodology allows us to work in this setting and outperform state-of-the-art methods under the same conditions (shown in our results). 
We address the GZSL task under this setting using $S^{lim\_sup} = \{S^{seen} \bigcup S^{AUD}\}$ to train our model. For this setting, $S^{AUD}$ includes unlabeled samples from seen classes only (in contrast to learn-from-anywhere setting where \emph{out-of-data} samples/classes were also a part of $S^{AUD}$). The final performance is studied on a test set that includes class labels from $Y^{seen} \bigcup Y^{novel}$.

\vspace{-4pt}
\subsection{Recognition in Test Phase}
\vspace{-4pt}
 We use the POE network to compute the joint representations for each data sample. For paired seen class data, we use $\mu$ (output of Algorithm 1), i.e., the joint mean of different modalities. In the case of unpaired AUD samples or unseen classes, we simply use the mean vector (instead of the joint mean) since only image data (for AUD) or attribute data (for unseen classes) is available at training time. Next, we use these representations to train a single-layer feed-forward neural classifier on the seen classes (with 100 neurons in the hidden layer). 
 \paragraph{Inference:} Finally, the testing data samples (both seen and unseen) are transformed into joint representations and classified using trained classifier network ( as described above). The testing protocol is similar to the GZSL classification setup of CADA-VAE \cite{cada}.
\vspace{-4pt}
\section{Experiments and Results} \label{exp}
\vspace{-4pt}
We evaluate the performance of the proposed model on both  Generalized Zero-Shot Learning (GZSL) as well as Generalized Few-Shot Learning (GFSL) on four benchmark datasets: Caltech-UCSD-Birds (CUB), Scene classification with attributes (SUN), Animals with Attributes 1 and 2 (AWA1 and AWA2). We use the standard 312-dimensional attributes for CUB \cite{cub}; 85-dimensional attributes for AWA1 and AWA2 \cite{good_bad}; and 102-dimensional attributes for the SUN dataset \cite{sun}.\\ 

\begin{figure*}[h!]
\vspace{-4pt}
\centering\includegraphics[width=0.95\textwidth,height=0.17\textheight]{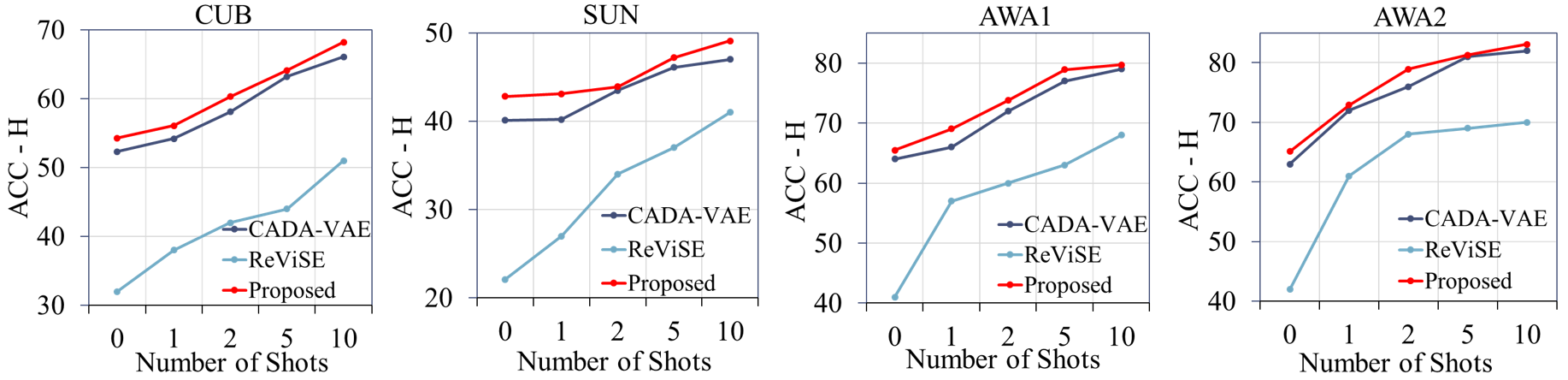}
\caption{Results on generalized few-shot learning with number of samples from unseen classes as 0, 1, 2, 5, and 10. }
\label{fig:fewshot}
\vspace{-4pt}
\end{figure*}

For evaluation across all four datasets, we use unlabeled images of \emph{out-of-data} classes from ImageNet and OpenImage \cite{OpenImages} as AUDs. We take 500 classes from both datasets with 500 images in each class. For ImageNet as well as OpenImage, we follow the general idea of a split from \cite{good_bad}; that is, the AUDs do not contain samples of any unseen classes, for fair evaluation. Furthermore, in the case where we use pseudo-auxiliary semantic embeddings, we extract 512-dimensional features 
from the image-to-text model \cite{botton_up_atten} trained on the MSCOCO dataset  (Note that we remove all unseen class samples beforehand so we do not violate the zero-shot condition)
We present the implementation details and time/space complexity information in the supplementary material due to space constraint.

\vspace{-4pt}
\subsection{Results: Learn from anywhere}
\vspace{-4pt}

In this section we discuss the performance of our proposed methodology on Learn from Anywhere paradigm. Note that the goal here is not to claim state-of-art performance but to demonstrate that by utilising abundantly available \emph{out-of-data} samples the proposed method helps to improve performance in generalized zero and few-shot settings.\par
\paragraph{Generalized Zero-Shot Learning:}
The  performance evaluation of the proposed methodology and comparison with several recently proposed GZSL methods is shown in Table \ref{tab:gzsl}. Note that we show results of our proposed approach for the cases where $S^{AUD}$ belong to \emph{out-of-data} classes from ImageNet and OpenImage datasets.
The H-score (harmonic mean of accuracy on seen and unseen classes) achieved by the proposed model on CUB and SUN is 56.1 and 43.9, respectively, higher than all the compared models. On the other hand, the proposed model achieves the H-score of 64.9 and 65.6 on AWA1 and AWA2, respectively, comparable to the best performing model. It can be clearly seen our method improves generalization to both seen and unseen classes at test-time and consistently performs better than the other methods on all the datasets.
\paragraph{Generalized Few-Shot Learning:} \label{GFSL}
\vspace{-4pt}
We compare the results of our proposed approach with two important image-semantic (pair) alignment based methods i.e CADA-VAE \cite{cada} and ReViSE \cite{revise} following the comparison in \cite{cada}. The results of generalized few-shot learning are presented in Figure \ref{fig:fewshot}. We vary the number of samples from unseen classes from 0 to 10, where 0 stands for zero-shot setting while the rest correspond to $k$-shot settings \cite{cada}. Expectedly, all methods perform better in GFSL than in GZSL since paired data of few-shot classes is present during the training. 
We notice that the proposed model significantly improves the performance over ReViSE  as well as  consistently outperforms CADA-VAE across all the considered datasets. 

\begin{figure}[h!]
\label{fig:weak}

\centering\includegraphics[scale=0.32]{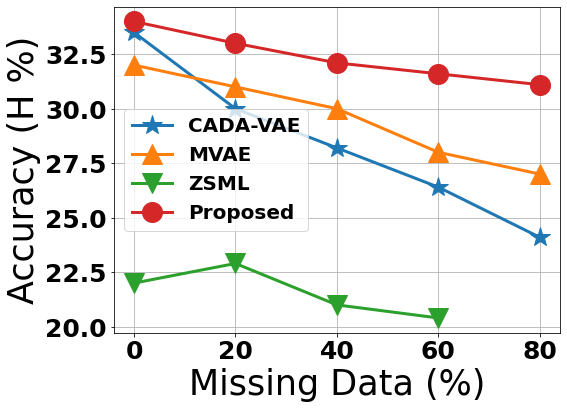}
\caption{GZSL with limited supervision. Performance of various models on CUB with a fraction of paired samples missing.}
\label{fig:weak}
\end{figure}

\vspace{-2pt}
\subsection{Results: GZSL with Limited Supervision} \label{gzsl_lim_sup}
\vspace{-4pt}
Using AUDs in our method allows us to function even when there is limited labeled data in the seen classes or the seen class samples have missing modalities.
To study the proposed model on GZSL with Limited Supervision, we gradually remove semantic information and labels from a  fraction of training data (seen classes) in our training procedure. We use the word vector encoding as the semantic information for the CUB dataset in this study, as provided in \cite{cada}.
We compare the proposed method with well-known recent methods, CADA-VAE \cite{cada}, ZSML \cite{verma2020meta}, and MVAE \cite{mvae} on this setting. 
We drop labels and semantic information at random from a percentage of training samples. In contrast, the novel class samples of test set are left as it is for a fair evaluation. For alignment methods such as CADA-VAE, which do not allow the use of unpaired image-semantic data, we remove the entire training sample (i.e image, semantic embedding pair). On the other hand, since our method can operate with unpaired data, the samples with missing semantic embedding are treated as AUDs in our case. The results are presented in Figure \ref{fig:weak}. It can be clearly seen that as the fraction of samples with missing semantic information and labels increases,  the proposed method outperforms all other methods by a considerable margin while giving a consistent performance - even when 80$\%$ of the auxiliary data is unavailable, showing it's robustness under such scenarios. The performance of CADA-VAE  decreases more quickly than others, while MVAE and the proposed method achieve consistent performance because these methods use POEs to model the joint distribution. 
(Note that performance at 0\% missing data does not match Table 1 since we use word vectors as auxiliary information for this analysis following \cite{cada}, instead of attributes).


\vspace{-4pt}
\subsection{Results: Large-scale Experiments} \label{imagenet_exp}
\vspace{-4pt}
Here, we evaluate the proposed method on ImageNet, which is a challenging GZSL dataset. We use the eight splits provided by \cite{good_bad} for the GZSL setup in this regard. The first two splits, $1H$ and $2H$, denotes all classes that are 2-hops and 3-hops away from the original $1K$ classes according to the ImageNet label hierarchy. These two splits evaluate the proposed method for generalization on hierarchical or semantic similarity among classes. The other six splits evaluate the proposed model on highly imbalanced classes with $M500, M1K, M5K$ being the most populated classes while $L500, L1K, L5K$ being the least populated comes from the remaining $21K$ classes. Finally, the $all$-split contains all classes. As the class-attributes are not available for ImageNet, we use word2vec embeddings given by \cite{sync} as semantic representation, and the ResNet-101 visual features are taken from \cite{good_bad} (we use the same splits and features as provided by \cite{good_bad} for fair Comparison). The AUD is constructed from OpenImages \cite{OpenImages}, where we remove the data corresponding to zero-shot classes. For this experiment, we extract pseudo-auxiliary semantic embeddings corresponding to AUD samples, as described in section \ref{learn-from-anywhere}. Note that for fair comparison, we make sure that unseen class samples are not used in AUDs or in any other way during training.

\begin{figure}[t!]
\label{fig:gfsl}
\centering\includegraphics[width=0.46\textwidth,height=0.15\textheight]{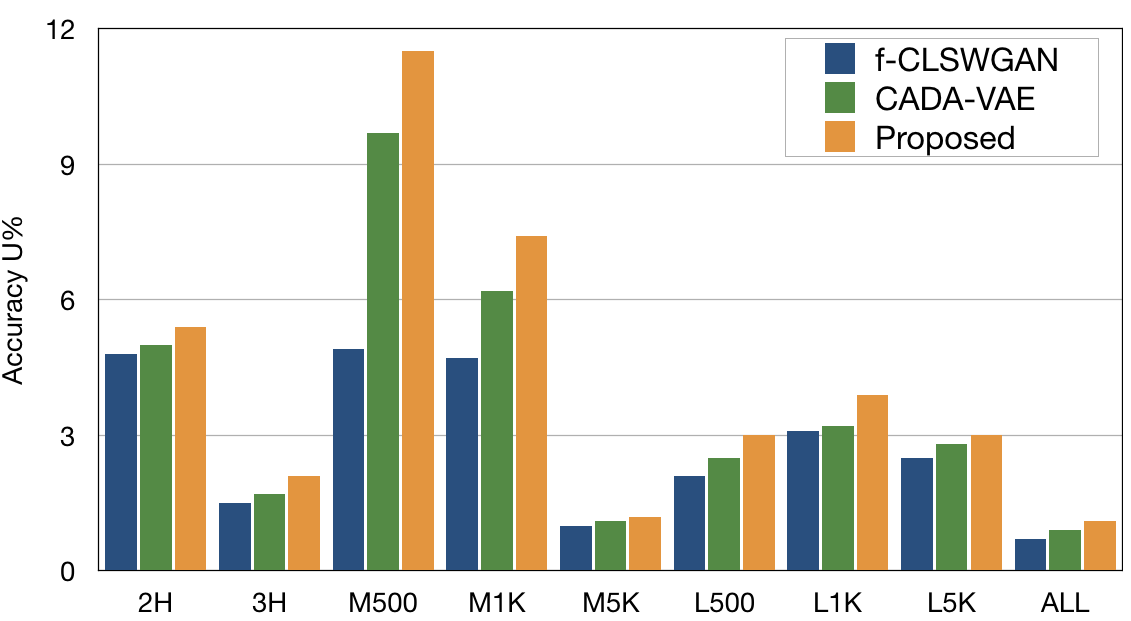}
\caption{GZSL results on ImageNet with OpenImages as AUD}
\label{fig:imagenet}
\end{figure}


Figure \ref{fig:imagenet} shows results for ImageNet dataset. The proposed method performs significantly better than f-CLSWGAN \cite{f-clswgan} and CADA-VAE \cite{cada} baselines. More populated class accuracy is expectedly higher than the least populated classes for 500 and 1K. With the addition of AUDs, the class bias is minimized, and the proposed model shows a significant increase in classification accuracy on 500 and 1K classes for both more and least populated labels. Furthemore, we notice that our proposed approach is able to get better performance on all scenarios  considered for the ImageNet experiment.

\section{Ablation Study} \label{ablation}
\vspace{-4pt}
We conducted ablation studies of the various components in our framework on the CUB dataset - in particular, by evaluating the effect of the use of skip connections, AUDs, and pseudo-auxiliary semantic embedding. Table \ref{tab:ablate} shows these results, along with the performance of MVAE \cite{mvae} for comparison purposes. 
The second row of Table \ref{tab:ablate} is the proposed model without skip-connections, whereas all the other variants of the proposed architecture have skip-connections. The proposed model without skip-connection and AUDs should behave like MVAE (in theory); however, we found that Swish activation used in MVAE architecture significantly degrades the GZSL setup's performance. With the addition of skip-connections and ReLU activation, the proposed model performs significantly better in terms of classification accuracy on seen and unseen classes.
The H-score without AUDs (with skip connection, third-row Table \ref{tab:ablate}) is 51.7, which is significantly higher than the MVAE model - showing the importance of skip-connections. On introducing AUDs, the H-score increases to 53.4, increasing both seen and unseen class accuracies. Finally, pseudo-auxiliary semantic information further increases performance across all metrics. 


\noindent \textbf{Effect of using POE network}:
We can follow two design choices for sampling the latent variable $z$: (i) Using the POE networks; and (ii) Multiplying latent variables corresponding to each modality. The ablation results of both these design choices are shown in Figure \ref{fig:z}. The architecture with the latent variable sampled from POEs results in significantly higher classification accuracy with each epoch. This also shows that POEs are better suited to the proposed architecture.

\begin{figure}[t!]
\setlength\belowcaptionskip{-10pt}
\label{fig:z}
\centering\includegraphics[width=0.45\textwidth,height=0.17\textheight]{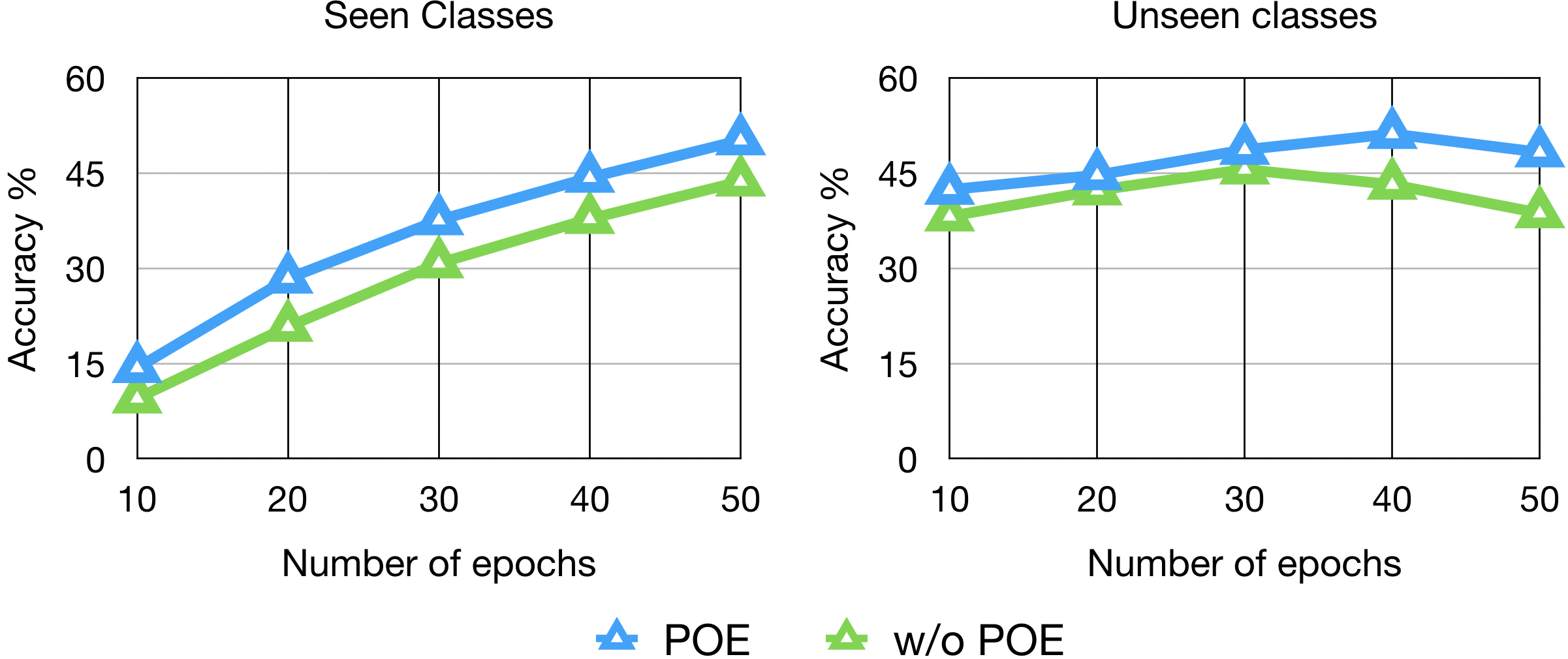}
\caption{Effect of using POE network 
}

\label{fig:z}
\end{figure}

\begin{table}[t!]
\centering
\resizebox{\columnwidth}{!}{
\begin{tabular}{llll}
\hline
Model  & S & U & H \\
\hline
MVAE  \cite{mvae} & 44.5 & 47.7 & 45.1 \\
proposed (w/o skip-connections) & 47.6 & 50.7 & 49.1 \\
proposed + skip-connections & 51.6 & 51.9 & 51.7 \\
proposed + AUDs & 54.2 & 52.8 & 53.4 \\
proposed + AUDs + pseudo-auxiliary semantics & 54.7 & 53.9 & 54.3 \\
\hline
\end{tabular}
}
\caption{Ablation study of the proposed model on CUB dataset}
\vspace{-15pt}
\label{tab:ablate}
\end{table}

\vspace{-14pt}
\section{Conclusion}
\vspace{-4pt}
In this work, we focused on improving generalized any-shot learning by using unannotated data, viz, unlabeled data without attribute information, which is not exploited generally in GZSL. The proposed method utilizes the unannotated data from various sources to reduce the bias towards seen classes in GZSL and GFSL. We demonstrate through various experiments on GZSL, GFSL, and GZSL with limited supervision on multiple benchmark datasets that the proposed technique has an advantage over existing state-of-the-art as it can leverage unannotated data in such settings and tackle the missing modality problem as well. The presented method is relatively general and can be used in any similar setting where the manual annotation is a bottleneck.

\bibliographystyle{named}
\bibliography{ijcai21}

\newpage

\appendix

\section{Supplementary Section}

In this Supplementary Material, we present some additional ablation studies that could not be included in the main paper due to space constraints:

\begin{itemize}
\setlength\itemsep{-0.2em}
    \item Training details of our experiments (in continuation to Section 4).
    \item Discussion related to time and space complexity of our approach (in continuation to Section 4).
    \item Visualizing samples from AUDs and corresponding pseudo-attribute embedding (in continuation to Section 3.2).
    \item Standard zero-shot learning experiments on CUB, SUN. AWA1 and AWA2 (in continuation to Section 4.1).
    \item Varying number of AUD samples (in continuation to Section 5).
    \item Design choice of pseudo-attribute embeddings (in continuation to Section 3.2).
    \item Varying the latent dimensions and AUD factor $\gamma$ (in continuation to Section 5).
\end{itemize}

\subsection{Training Details}
\vspace{-4pt}
We use a single-hidden-layer feedforward neural network with 1400 and 550 neurons for encoders and decoders, respectively.
For AUDs, we share the image encoder and image decoder parameters, while 200 neurons are used for the pseudo-auxiliary embedding decoder. We follow the evaluation setup of [23] and use ResNet-101 features as input to our model. 
For testing, we use the POE network to compute the joint representations for each data sample (seen/unseen test images), i.e., $\mu$. The joint representations are used to train a single layer feedforward neural classifier on the seen classes (with 100 neurons in the hidden layer). Finally, the testing data samples (both seen and unseen) are transformed into the joint representations and classified using this trained network. The testing protocol is similar to the GZSL classification setup of CADA-VAE [14].

\begin{table}[!b]
\footnotesize
\centering
 
\vspace{-3pt}
\label{tab:gzsl}
\begin{tabular}{|c|c|c|c|}
\hline
\textbf{Dataset}     & \multicolumn{1}{c|}{\textbf{CUB}} & \multicolumn{1}{c|}{\textbf{AWA2}} &
\multicolumn{1}{c|}{\textbf{SUN}}  \\ \hline
\textbf{Methods}     & T1     & T1         & T1             \\ \hline
\textbf{CONSE}(ICLR 2014)~               & 34.3        & 44.5        & 38.8             \\
\textbf{SSE}(ICCV 2015)~                 & 43.9        & 61.0        & 51.5          \\ 
\textbf{LATEM(}CVPR 2016)~               & 49.3        & 55.8        & 55.3             \\ 
\textbf{ALE}(TPAMI 2016)~                 & 54.9        & 62.5        & 58.1          \\ 
\textbf{DEVISE}(NIPS 2013)~              & 52.0        & 59.7        & 56.5              \\ 
\textbf{SJE}(CVPR 2015)~                 & 53.9        & 61.9        & 53.7        \\ 
\textbf{ESZSL}(ICML 2015)~               & 53.9.       & 58.6        & 54.5     \\ 
\textbf{SYNC}(CVPR 2016)~                & 55.6        & 46.6        & 56.3         \\ 
\textbf{SAE}(CVPR 2017)~                 & 33.3        & 54.1        & 40.3            \\ 
\textbf{GFZSL}(ECML 2017)~               & 49.2        & 67.0         & 62.6            \\ 
\textbf{CVAE-ZSL}(CVPRW 2018)~            & 52.1        & 65.8         & 61.7          \\ 
\textbf{SE-ZSL}(CVPR 2018)~              & 59.6        & 69.2         & 63.4             \\ 
\textbf{DCN}(NIPS 2018)~                 &56.2         & -            & 61.8             \\ 
\textbf{JGM-ZSL}(ECCV 2018)~             &54.9         & 69.5         & 59.0            \\ 
\textbf{RAS+cGAN}(NC 2019)~            &52.6         & -            & 61.7         \\ 

\textbf{DEM}(CVPR 2017)~                 &51.7         & 67.1         & 61.9        \\ 
\textbf{SP-AEN}(CVPR 2018)~             &55.4         & 58.5         & 59.2         \\ 
\textbf{f-clsWGAN}(CVPR 2018)~             &57.3         & 68.2         & 60.8        \\ 
\textbf{CADA-VAE}(CVPR 2019)~             &60.4         & 64         & 61.8         \\ 
\textbf{f-VAEGAN}(CVPR 2019)~             &61.0         & 71.1         & 64.7        \\ 
\textbf{GZLOCD}(CVPR 2020)~             &60.3         & 71.3         &63.5       \\ \hline

\textbf{Proposed+AUDs} (Imagenet) & \textbf{66.1}  & \textbf{75.7} & \textbf{65.5}\\ 
\textbf{Proposed+AUDs} (OpenImages) & \textbf{68.5}  & \textbf{76.5} & \textbf{66.8}\\ \hline


\end{tabular}
\caption{Standard ZSL results on CUB, SUN and AWA2. Here, we report top-1 (T1) accuracy on all the datasets.}
\vspace{-9pt}
\label{tab:zsl}
\end{table}

\begin{figure*}
    \centering
    \setlength\belowcaptionskip{-18pt}
    \includegraphics[scale=0.43]{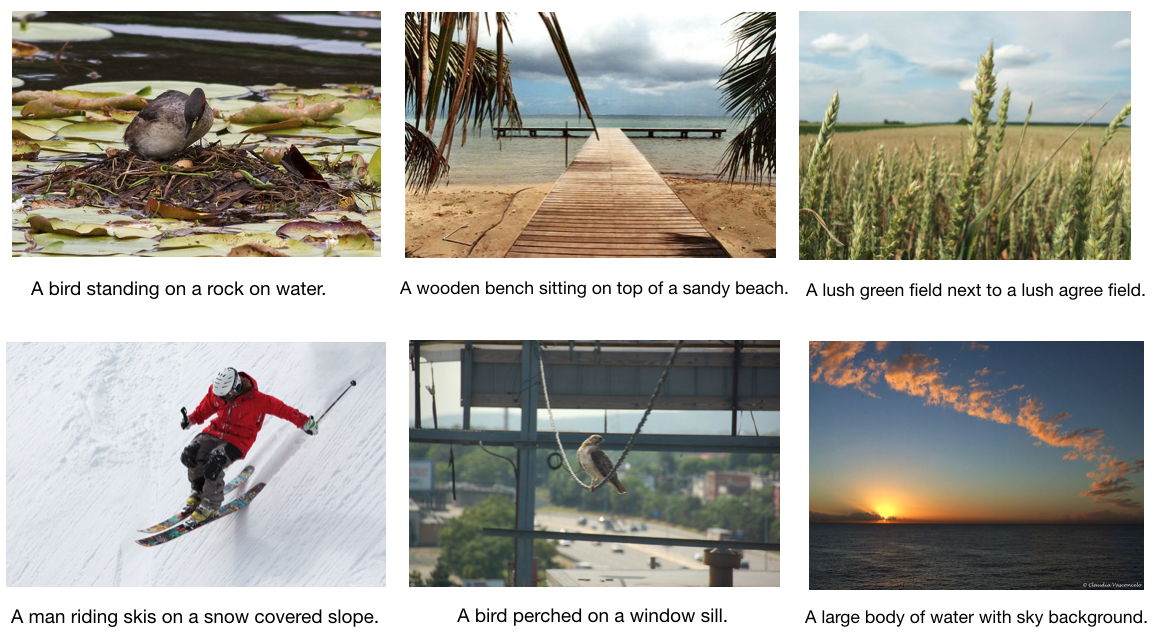}
    \caption{Visualizing samples from AUDs (ImageNet). Here, the pseudo-attributes (sentences) generated by image-to-text generator. Note that the images are unannotated and no corresponding label is provided during training.}
    \label{fig:dec}
\end{figure*}

We use a batch size of 32 across all datasets. The size of the latent embedding that we use is 128. We compute the KL-divergence term for joint computation using annealing technique, where the weight $\beta_{i} (i \in \{image, text, AUD\})$ of KL-term is increased by a rate of 0.0035 per epoch until 85. We use the annealing strategy for $\gamma\ $, where $\gamma$ is increased from epoch 10 to 56 by a factor of 0.005 per epoch. 
The value of $ \alpha $ is taken as 0 or 1 (0 for AUD and 1 otherwise). 

\subsection{Time and Space Complexity}
\vspace{-4pt}
The AUD dataset is around the same size as the training set. The total number of samples that the model encounters during training (AUD+seen) is 2-3 times the seen class samples. Given that our method requires ResNet-101 features, the increased number of training samples does not pose much difference in training time. We also observed that increasing the batch size from 32 to 48 takes the same amount of time as the standard ZSL task, without affecting the ZSL performance.

\subsection{Standard Zero-shot Learning Results}
\vspace{-4pt}

We present the standard ZSL results here. The results are shown in Table \ref{tab:zsl}, where we experiment with both ImageNet and OpenImages AUD samples. In order to ensure an exhaustive comparison, we compare with all state-of-the art ZSL methods, including recent ones, as mentioned in the very recent work [5]. Furthermore, we also compare with some other important ZSL methods like f-VAEGAN, CADA-VAE, f-CLSWGAN. It can be clearly seen that even on the standard ZSL setting, our method outperforms other methods (including ones specifically designed explicitly for this setting) on CUB, AWA2 and SUN datasets. It should be noted that the choice of AUDs also affect the overall performance. We observe that OpenImages have slight better pseudo-attribute generated than the ImageNet, hence the performance is better for OpenImages.



\subsection{Varying Number of AUD Samples}
\vspace{-4pt}
Here, we study the effect the varying the number of samples in AUDs. Since our goal is to minimize the class bias of seen classes and improve classification performance on the unseen classes, the choice of AUDs should make a difference. We speculate that the choice of AUD is more critical than the number of samples. To verify this, we perform an ablation study by varying the number of samples of ImageNet from 1K to 100K to see the effect of classification on the CUB dataset. We present two separate runs of experiments where the number of AUDs is chosen randomly. Thus the two runs differ only in the quality of AUDs and not quantity. 

\begin{figure}[h!]
\label{fig:gfsl}

\centering\includegraphics[scale=0.2]{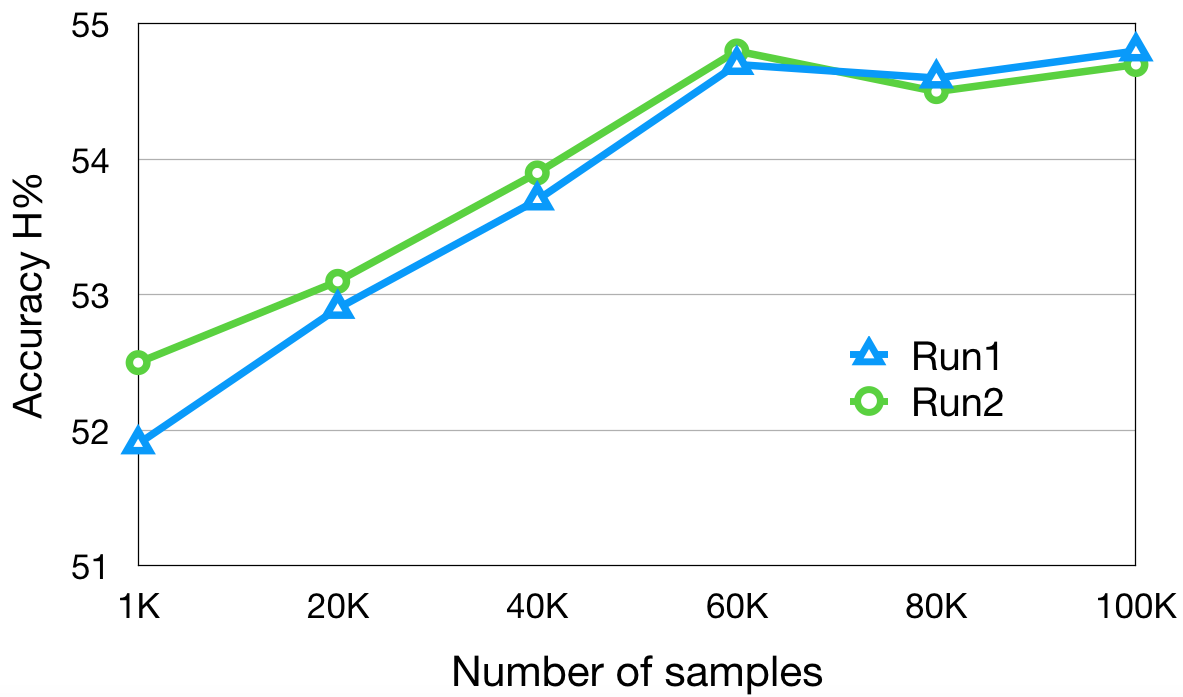}
\caption{Performance on varying the number of AUD samples}
\label{fig:inc_sam}
\end{figure}

The results are shown in Figure \ref{fig:inc_sam}. For the addition of 1K AUDs, the classification performance is not the same for both runs. The set of AUDs in $Run1$ is not as relevant as $Run2$. However, after the addition of 50K samples, the performance of both models is identical. With a large set, the chances of getting relevant AUDs are high. We also note that using a huge AUD set is also not desirable, as the computational cost increases with the addition of AUDs. Keeping the AUDs set to be approximately two times the training set works well for all of our experiments.

\subsection{Pseudo-attribute Embedding}
\vspace{-4pt}
In Section 3, we stated that pseudo-attributes are generated using an image-to-text generative model. We can also generate the captions corresponding to the unannotated image followed by word-vector synthesis for the generated captions using a pre-trained Word2vec model. Here, we present an ablation analysis over both the design choices on the CUB dataset with ImageNet as AUD.
For the first design choice, we use the 512-dimensional penultimate layer features of the pre-trained image-to-text model, while for the second design choice, we compute the captions corresponding to the AUDs. The vector embedding is computed as a sum of vectors of each word embedding:
\begin{align}
a(\Tilde{x}) = \sum_{t}^T w2v(t),
\end{align}
where T is the total number of words generated for the given image, and w2v is the pre-trained word2vec model \footnote{https://code.google.com/archive/p/word2vec/}. 

These results are shown in Table \ref{tab:ablate_wv}. Both design choices result in a similar performance on the CUB dataset, although the use of image-to-text was marginally better in this case. In general, in our work, we found no significant difference between these design choices, and one could use one of them based on their availability in a newer setting.

\begin{table}[h!]
\centering
\begin{tabular}{llll}
\hline
Model  & S & U & H \\
\hline
image-to-text embedding & 56.8 & 52.1 & 54.3 \\
word2vec embedding & 55.4 & 52.4 & 53.8 \\
\hline
\end{tabular}
\caption{Performance with difference pseudo-attribute embedding design choices on CUB dataset}
\label{tab:ablate_wv}
\end{table}

\subsection{Varying $\gamma$ and Size of Latent Dimensions}
\begin{figure}[t!]
\label{fig:gfsl}

\centering\includegraphics[width=0.5\textwidth,height=0.17\textheight]{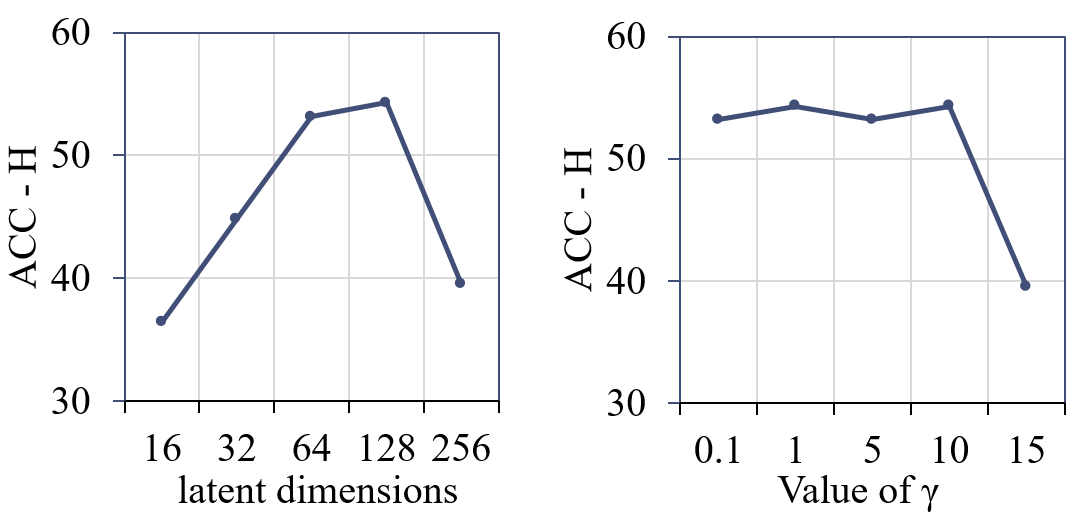}
\caption{Performance on varying the number of latent dimensions and $\gamma$}
\label{fig:gamma_lat}
\end{figure}

\vspace{-4pt}
We also studied various values of $\gamma$ (pseudo-auxiliary embedding factor) and different latent dimensions. The experiments are conducted on the CUB dataset, and the results are shown in Figure \ref{fig:gamma_lat}. For fewer latent dimensions (16 and 32), the models achieve a low H-score. Similarly, for high values (256 and beyond), the performance degrades. With fewer latent dimensions, the model does not have enough capacity to capture the entire latent space, while with very high dimensions, the architecture suffers from instability during optimization. 

For a minimal value for the pseudo-auxiliary factor (0.01 and below), the importance given to the pseudo-auxiliary decoder is negligible, and the performance is equivalent to the proposed+AUDs. When $\gamma$ is increased to 5 or beyond, the performance suddenly drops down. With very high values of $\gamma$, the model is biased towards the word-vector embedding. Since the word-vector embedding results in lower performance than the attribute embedding [14], this drop in H-score is not surprising. Another reason for the drop in performance can be the quality of pseudo-auxiliary embedding. Not all attributes generated for unannotated data may be relevant, and thus with high values of $\gamma$, the model is biased towards irrelevant samples.

\end{document}